# A Topic-aware Comparable Corpus of Chinese Variations

Da-Chen Lian, Shu-Kai Hsieh
Graduate Institute of Linguistics
National Taiwan University
r05142009@gmail.com, shukaihsieh@ntu.edu.tw

**Abstract**

This study aims to fill the gap by constructing a topic-aware comparable corpus of Mainland Chinese Mandarin and Taiwanese Mandarin from the social media in Mainland China and Taiwan, respectively. Using Dcard for Taiwanese Mandarin and Sina Weibo for Mainland Chinese, we create a comparable corpus that updates regularly and reflects modern language use on social media.

## 1. Introduction

With its economic and cultural significance, the notion of 'World Chineses' has been widely recognized in recent years, and the studies of variation of World Chineses at different linguistic levels (e.g., lexical and grammatical) are beginning to unfold (Lin et al, 2018).

However, comprehensive empirical research on variations as well as computational linguistic applications are hindered by the lack of availability of dynamically updated comparable corpora of different varieties of Mandarin Chinese.

The remainder of this paper is structured as follows: Section 2 introduces background knowledge of comparable corpus and our proposed approach. In Section 3, we describe the compilation process of the corpus. Section 4 illustrates an example based on the corpus, and Section 5 concludes the paper.

## 2. Comparable Corpus and Language Variations

*Comparable Corpus,* commonly defined as a collection of 'similar' texts in different languages or in different varieties of a language, has been widely recognized as an essential language resource especially for contrastive linguistic or translation studies. Though terminological variations remain, a related but distinguishable type of corpus is called Parallel Corpus, which mainly consists of a set of texts in one language and its translations in other language(s). It is noted that what links the collections of (independent) texts in comparable corpora is that they are *aligned* according to the same criteria, such as texts on a certain topic, from a given context, from a certain period, etc (Kenning, 2010). Parallel corpora are useful 'for exploring how the same content is expressed in two languages' (Aijmer & Altenberg, 1996, p. 13, as cited in McEnery & Xiao, 2007, p. 21). Even so, they still suffer from the effect of being translations and may not be suitable for the study of cross-linguistic contrasts (ibid). In contrast, "comparable corpora are more useful for translations studies" with several studies indicating translators increase their productivity and improve their quality with access to an appropriate comparable corpus (Friedbichler & Friedbichler, 1997, as cited in McEnery & Xiao, 2007). The most significant advantage in using large-scale comparable corpora "is that it enables the comparison of different languages or varieties in 'similar circumstance of communication, but avoiding the inevitable distortion introduced by the translations of a parallel corpus (Eagles, 1996, cited in Huang et al, 2011, p. 399).'" Several more well-known Mandarin Chinese corpora include the Chinese Gigaword corpus, the LIVAC (Linguistic Variations in Chinese Speech Communities), and the PKU-CCL corpus (Huang et al, 2011).

Though there is an array of studies dedicated to the language variations and cross-language translational equivalents based on parallel corpora, available comparable corpora in Chinese are not so easily found.
An example of language variation that can be gathered from using a large comparable corpus is how the light verb 進行 *jinxing* 'proceed' is used differently in Mainland Mandarin and Taiwanese Mandarin (ibid).

## 3. Corpus Compilation

Previous studies have mostly focused on linguistic phenomenon. The purpose of this project is on creating a resource. This resource is meant to reflect more modern use of language through text gathered on social media on a consistent basis. In addition to its size and dynamic nature in longitudinal Web-as-Corpus sampling, this corpus features three aspects: (1) short-text oriented, (2) Hashtag-as-common topic, and (3) Machine alignment.

### 3.1 Data Ingestion

Dcard (www.dcard.tw) was chosen because while users of the website do not use hashtags explicitly, users can provide tags for every post they make. These tags serve as metadata, which serve to group posts that share the same tags together. This essentially serves a similar purpose as hashtags found in a post itself.

Dcard is a popular forum in Taiwan mainly targeted towards university students. This website is separated into different boards depending on the topic. Boards also exist for a plethora of universities found in Taiwan. In total, there are 236 different boards. According to an online competitive intelligence tool that tracks various factors of websites, *SimilarWeb*, as of this writing, Dcard is ranked 41st in Taiwan in overall traffic and 2843rd globally with 18.69 million visits in the last six months.

Sina Weibo (www.weibo.com) is a very popular microblogging site in mainland China that can be considered very similar to Twitter. It was chosen because of the prevalence of hashtags found on the social website. Unlike Dcard, there are no established topics that users must choose under which to post. A character limit of 140 is also imposed. According to the aforementioned intelligence tool, Weibo ranks 20th in mainland China and 111st globally. In the last six months, 303.66 million people have visited the microblogging site.

Texts as well as metadata from Dcard are crawled via an available API. Due to the targeted demographic for this website, metadata regarding users' place of study are also available. Metadata includes the gender of the user, where s/he studies, his or her department, user-given tags associated with the post, the number of likes for the post, and the number of comments.

The API for Sina Weibo is much more restrictive on what can be collected, namely only one's own timeline, which is essentially the front page of one's own account where posts from accounts that one follows appear. Furthermore, only the first 50 pages is available to the average user.

Meta-information available include the number of followers of a user, the user's gender, the user's screenname, and the time a post was created.

### 3.2 Pre-processing

For the Dcard posts, the API provided made pre-processing very simple. All posts were presented in JSON format. Tags have its own key-value pair. Python regular expressions were used to remove URLs from posts. A customized version of the popular Chinese segmentation package *jieba* aimed towards Taiwanese Mandarin, aptly called *jieba-tw,* was used.

For the Weibo posts, the posts needed to be acquired manually without the assistance of an API. As such, much more pre-processing was needed. The Python package *BeautifulSoup* was used to extract only the text of each post since the majority of posts were replete with HTML. To extract hashtags and remove URLs, regular expressions were employed. Any string of text between two "#" symbols was considered a hashtag. For segmentation, the original *jieba* package was used. This is optimized for simplified Chinese.

Because of the manual nature of the collection process and the restrictions imposed on the website itself, the number of unique posts gathered was very small.

### 3.3 Machine alignment

Because there are many more Dcard posts compared to Weibo posts, the former are used as the basis on which to find a similar document from the smaller pool of Weibo posts. After the initial pooling of sentences based on hashtags, the Dcard posts go through a series of transformations via the topic modeling package *Gensim* (Rehurek & Sojka, 2010). First, each post is converted into a term frequency-inverse document frequency vector. Then, the tf-idf transformed corpus is again transformed into a 300-dimensional latent space via Latent Semantic Indexing. To align a Weibo post with Dcard posts, cosine similarity is used. The results are returned to the user as a list of the most similar documents in decreasing order.

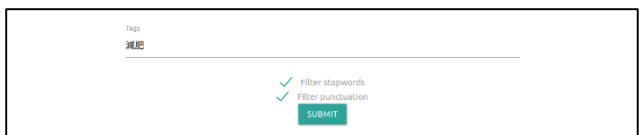

Image 1. Search box.

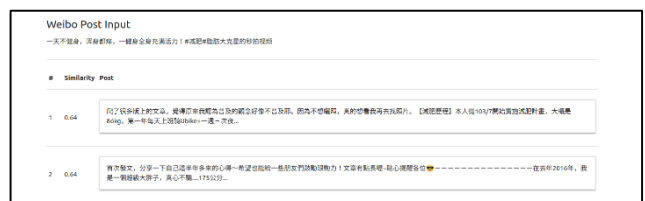

Image 2. Posts sorted by similarity.

### 3.4 Analysis Functions

At the top of the results page, quick statistics are provided for the posts, such as total posts from each respective social media source, the number of posts from women and men, the average post length, and the polarity on average for a particular tag in a specific website. Polarity is computed using a sentiment dictionary.

Next, given a tag as input, a random post from Weibo that contains the indicated tag will be chosen while a list of posts from Dcard that also contain the tag will be collected and then returned in a list sorted by highest similarity.

Further down, a frequency list for both websites sorted in descending order are presented for the posts found matching the query.

Collocations for the input are also provided with the different collocations from each social network presented side-by-side. We use NLTK's (Bird & Loper, 2004) *BigramCollocationFinder* class to search for bigram collocations in each respective pool of posts. Collocation strength is determined by pointwise mutual information (PMI).

Image 3. Collocations for 減肥 from Dcard

# 4. An Example

This section offers a concrete example of the ways in which the contrastive language uses can be analyzed with the aid of this corpus.

Take workout language as an example. As a proof of concept, three keywords, or tags, were chosen, somewhat arbitrarily, and used to gather relevant posts from both websites. These include 減肥 *jianfei* 'to lose weight', 健身 *jianshen* 'to workout', and 瘦身 *shoushen* 'to lose weight'. These three were chosen because of their similar meanings and because a topic with fewer images and videos and more text would be more conducive to returning meaningful results.

First, we will examine the results when the tag 健身 *jianshen* 'to workout' is queried. In Table 1, an overview of the search results can be seen. Immediately, we can see a discrepancy between the number of women and men who use the queried word. Dcard male users use the word a lot more than their female counterparts. The opposite can be said of Weibo users. Male users of the site use it less compared to its female users. It can also be seen that the use of the word is associated more positively when used in Dcard than in Weibo.

| Site | Total Posts | Men | Women | Avg. Post Length | Naïve Polarity |
|---|---|---|---|---|---|
| Dcard | 623 | 462 | 161 | 128.8 | 0.0481 |
| Weibo | 725 | 267 | 458 | 51.8 | 0.0157 |

Table 1. A quick overview of search results when 健身 *jianshen* 'to workout' is queried.

| | Dcard | Weibo |
|---|---|---|
| 1. | 會 *hui* 'can' | 健身 *jianshen* 'to workout' |
| 2. | 健身 *jianshen* 'to workout' | 动 *dong* 'to move' |
| 3. | 想 *xiang* 'to want' | 视 *shi* 'to look at' |
| 4. | 運動 *yundong* 'to exercise' | 頻 *pin* 'repeatedly' |
| 5. | 訓練 *xunlian* 'to train' | 秒 *miao* 'second' |

Table 2. Five most frequent tokens used when 健身 *jianshen* 'to workout' is queried.

| | Dcard | Weibo |
|---|---|---|
| 1. | 會觸發 *chufa* 'will set off' | 些人 *xieren* 會 'some people will' |
| 2. | 會變笨 *bianben* 'will become stupid' | 会引起 *inqi* 'will bring about' |
| 3. | 會促使 *cushi* 'will induce' | 会交流 *jiaoliu* 'to exchange' |

Table 3. Top 3 collocates for 會 *hui* 'will'.

Next, we will now look at the results when 減肥 *jianfei* 'to lose weight' is queried.

| Site | Total Posts | Men | Women | Avg. Post Length | Naïve Polarity |
|---|---|---|---|---|---|
| Dcard | 886 | 278 | 608 | 135.1 | 0.0345 |
| Weibo | 897 | 214 | 683 | 55.7 | 0.0135 |

Table 4. A quick overview of search results when 減肥 *jianfei* 'to lose weight' is queried.

While more male Dcard users than female ones use 健身 *jianshen* 'to lose weight' and more Weibo female users than male users use the same word, according to Table 4, there is a more consistent and clear divide in usage between male and female usage of the word on both websites.

| | Dcard | Weibo |
|---|---|---|
| 1. | 減肥 *jianfei* 'to lose weight' | 減肥 *jianfei* 'to lose weight' |
| 2. | 吃 *chi* 'to eat' | 瘦身 *shoushen* 'to lose weight' |
| 3. | 會 *hui* 'can' | 视 *shi* 'to look at' |
| 4. | 說 *shuo* 'to speak' | 頻 *pin* 'repeatedly' |
| 5. | 運動 *yundong* 'to exercise' | 秒 *miao* 'second' |

Table 5. Five most frequent tokens used when 減肥 *jianfei* 'to workout' is queried.

| | Dcard | Weibo |
|---|---|---|
| 1. | 吃得少 *deshao* 'eat little' | 吃货 *huo* 'foodie' |
| 2. | 吃手手 *shoushou* 'eat hands' | 忍不住 *renbuzhu* 吃 'cannot resist eating' |
| 3. | 吃土了 *tule* 'eat dirt' | 吃饱 *bao* 'eat until full' |

Table 6. Top 3 collocates for 吃 *chi* 'to eat'.

Finally, let us look at the results for 瘦身 *shoushen* 'to lose weight'.

| Site | Total Posts | Men | Women | Avg. Post Length | Naïve Polarity |
|---|---|---|---|---|---|
| Dcard | 164 | 65 | 99 | 158.5 | 0.0413 |
| Weibo | 913 | 139 | 774 | 48.8 | 0.0155 |

Table 7. A quick overview of search results of 瘦身 *shoushen* 'to lose weight' when queried.

Immediately obvious is the fact that this term is not widely used in Taiwanese Mandarin. Even so, there still appears to be a difference in usage between men and women.

| | Dcard | Weibo |
|---|---|---|
| 1. | 吃 *chi* 'to eat' | 瘦身 *shoushen* 'to lose weight' |
| 2. | 運動 *yundong* 'to exercise' | 減肥 *jianfei* 'to lose weight' |
| 3. | 會 *hui* 'can' | 视 *shi* 'to look at' |
| 4. | 瘦 *shou* 'skinny' | 頻 *pin* 'repeatedly' |
| 5. | 瘦身 *shoushen* 'to lose weight' | 动 *dong* 'to move' |

Table 8. Five most frequent tokens of 瘦身 *shoushen* 'to lose weight'.

| | Dcard | Weibo |
|---|---|---|
| 1. | 瘦到奶 *daonai* 'breasts lose weight' | 瘦大腿 *datui* 'lose weight in the thighs' |
| 2. | 瘦得快 *dekuai* 'lose weight quickly' | 瘦下半身 *xiabanshen* 'lose weight in the lower half of the body' |
| 3. | 瘦下來 *xialai* 'to lose weight' | 安心 *anxin* 瘦 'lose weight with peace of mind' |

Table 9. Top 3 collocates of 瘦 *shou* 'skinny'.

| Dcard | | | Weibo | | |
|---|---|---|---|---|---|
| # | Token | Frequency | # | Token | Frequency |
| 1 | 減肥 | 1193 | 1 | 減 | 1518 |
| 2 | 吃 | 1144 | 2 | 肥 | 1284 |
| 3 | 會 | 757 | 3 | 瘦身 | 338 |
| 4 | 就 | 707 | 4 | 視 | 330 |
| 5 | 運動 | 703 | 5 | 胸 | 323 |
| 6 | 想 | 603 | 6 | 吃 | 294 |

Image 4. Frequency list for the token 減肥 *jianfei* 'to lose weight'.

## 5. Conclusion

We present a topic-aware comparable corpus system for Chinese variations. Given a particular query, it returns Mainland Chinese-Taiwanese Mandarin short texts pairs that meet a certain similarity threshold. Along with some common corpus functions and NLP techniques, language usage at different levels with various meta-information can be provided for the comparison of Chinese language variations, and it also serves as training data for a short-text/phrase-level sequence to sequence neural network model for variation translation as well.

This can only be achieved if a sufficient amount of data can be gathered from Weibo or a similar website. As it stands, the data is simply insufficient to derive reliable conclusions. We leave the data enrichment and dynamic updating for future works.

## 6. Bibliographical References

APCLab. (2013-). Jieba-tw. https://github.com/APCLab/jieba-tw

Bird, S., & Loper, E. (2004, July). NLTK: the natural language toolkit. In *Proceedings of the ACL 2004 on Interactive poster and demonstration sessions* (p. 31). Association for Computational Linguistics.

Huang, C. R., Lin, J., & Zhang, H. (2011). World Chineses Based on Comparable Corpus: The Case of Grammatical Variations of jìnxíng. In *Sixth Cross-Strait Modern Chinese Symposium*.

Kenning, M. M. (2010). What are parallel and comparable corpora and how can we use them. *The Routledge handbook of corpus linguistics*, 487-500.

Lin, J., Shi, D., Jiang, M and Huang, C-R. (2018). Variations in World Chineses. In: Huang, Jing-schmidt and Meisterernst (eds). *The Routledge Handbook of Chinese Applied Linguistics*.

McEnery, T., & Xiao, R. (2007). Parallel and comparable corpora: What is happening. *Incorporating Corpora. The Linguist and the Translator*, 18-31.

Rehurek, R., & Sojka, P. (2010). Software framework for topic modelling with large corpora. *In In Proceedings of the LREC 2010 Workshop on New Challenges for NLP Frameworks.*

Richardson, L. (2014-). Beautiful Soup. https://www.crummy.com/software/BeautifulSoup/bs4/doc/

Sun J. (2013). Jieba. https://github.com/fxsjy/jieba